\documentclass{article}

\PassOptionsToPackage{numbers, compress}{natbib}
\newif\ifpreprintversion
\preprintversiontrue

\ifpreprintversion
  \usepackage[preprint]{neurips_2025}
\else
  \usepackage{neurips_2025}  
\fi

\usepackage{xcolor}         
\definecolor{linkColor}{rgb}{0.2,0.4,0.6}
\usepackage[utf8]{inputenc} 
\usepackage[T1]{fontenc}    
\usepackage[colorlinks=true,linkcolor=linkColor,citecolor=linkColor,filecolor=linkColor,urlcolor=linkColor]{hyperref}       
\usepackage{url}            
\usepackage{booktabs}       
\usepackage{amsfonts}       
\usepackage{nicefrac}       
\usepackage{microtype}      
\usepackage{xcolor}         
\usepackage{arydshln}

\RequirePackage{algorithm}
\RequirePackage{algorithmic}

\usepackage{multirow}
\usepackage{amsmath}
\usepackage{capt-of}
\usepackage{tabularx}
\usepackage{epsfig}
\usepackage{amssymb}
\usepackage{amsfonts}
\usepackage{booktabs}
\usepackage{scalerel}
\usepackage[inline]{enumitem}
\usepackage{listings}
\usepackage{varwidth}
\usepackage[export]{adjustbox}
\usepackage{tikz}
\usetikzlibrary{tikzmark}

\usepackage{stmaryrd}
\usepackage{bbm}
\usepackage{wrapfig}
\usepackage{pifont}
\usepackage[noabbrev]{cleveref}

\definecolor{deepblue}{rgb}{0,0,0.5}
\definecolor{officeblue}{RGB}{0,102,204}
\definecolor{deepred}{rgb}{0.6,0,0}
\definecolor{deepgreen}{rgb}{0,0.5,0}
\definecolor{mybrickred}{RGB}{182,50,28}

\definecolor{fillcolor}{RGB}{216,217,252}

\usepackage{etoolbox}
\usepackage{framed}

\newif\ifxetexorluatex
\ifxetex
  \xetexorluatextrue
\else
  \ifluatex
    \xetexorluatextrue
  \else
    \xetexorluatexfalse
  \fi
\fi
\newcommand*\quotesize{60} 
\newcommand*{\openquote}
   {\tikz[remember picture,overlay,xshift=-4ex,yshift=-2.5ex]
   \node (OQ) {\fontsize{\quotesize}{\quotesize}\selectfont``};\kern0pt}

\newcommand*{\closequote}[1]
  {\tikz[remember picture,overlay,xshift=4ex,yshift={#1}]
   \node (CQ) {\fontsize{\quotesize}{\quotesize}\selectfont''};}

\colorlet{shadecolor}{white}

\newcommand*\shadedauthorformat{\emph} 

\newcommand*\authoralign[1]{%
  \if#1l
    \def\authorfill{}\def\quotefill{\hfill}
  \else
    \if#1r
      \def\authorfill{\hfill}\def\quotefill{}
    \else
      \if#1c
        \gdef\authorfill{\hfill}\def\quotefill{\hfill}
      \else\typeout{Invalid option}
      \fi
    \fi
  \fi}
{\authoralign{#1}
\ifblank{#2}
   {\def\shadequoteauthor{}\def\yshift{-2ex}\def\quotefill{\hfill}}
   {\def\shadequoteauthor{\par\authorfill\shadedauthorformat{#2}}\def\yshift{2ex}}
\begin{snugshade}\begin{quote}\openquote}
{\shadequoteauthor\quotefill\closequote{\yshift}\end{quote}\end{snugshade}}

\newfloat{Code}{htbp}{Code}

\usepackage{amsmath,amsfonts,bm}

\def\eqref#1{equation~\ref{#1}}

\def\1{\bm{1}}

\DeclareMathAlphabet{\mathsfit}{\encodingdefault}{\sfdefault}{m}{sl}
\SetMathAlphabet{\mathsfit}{bold}{\encodingdefault}{\sfdefault}{bx}{n}

\newcommand\our{ReSA}

\title{Rectified Sparse Attention}

\author{
Yutao Sun\thanks{~Equal contribution. $\diamond$ Corresponding author.}$~~^{12}$~~~~~Tianzhu Ye\footnotemark[1]$~~^{12}$~~~~~Li Dong\footnotemark[1]$~~^{1}$~~~~~Yuqing Xia\footnotemark[1]$~~^{1}$ \\
~\bf Jian Chen$^{1}$~~~~~Yizhao Gao$^{13}$~~~~~Shijie Cao$^{1}$~~~~~Jianyong Wang$^{2}$~~~~~Furu Wei$^{1}$$^{\diamond}$ \\
~$^1$ Microsoft Research ~~~~~
~$^2$ Tsinghua University \\
~$^3$ The University of Hong Kong \\
~{\href{https://aka.ms/GeneralAI}{https://aka.ms/GeneralAI}}
}

\begin{document}

\maketitle

\begin{abstract}
Efficient long-sequence generation is a critical challenge for Large Language Models. While recent sparse decoding methods improve efficiency, they suffer from KV cache misalignment, where approximation errors accumulate and degrade generation quality. In this work, we propose Rectified Sparse Attention (\our{}), a simple yet effective method that combines block-sparse attention with periodic dense rectification. By refreshing the KV cache at fixed intervals using a dense forward pass, \our{} bounds error accumulation and preserves alignment with the pretraining distribution. Experiments across math reasoning, language modeling, and retrieval tasks demonstrate that \our{} achieves near-lossless generation quality with significantly improved efficiency. Notably, \our{} delivers up to 2.42$\times$ end-to-end speedup under decoding at 256K sequence length, making it a practical solution for scalable long-context inference.
\ifpreprintversion
Code is available at \url{https://aka.ms/ReSA-LM}.
\else

\fi
\end{abstract}

\section{Introduction}

The ability to process long contexts has become a core requirement for Large Language Models, with context lengths up to millions of tokens~\cite{gemini1.5,qwen2.5-1m}. In particular, long sequence generation has received growing attention, especially due to the demand for test-time scaling~\cite{deepseekr1, o1}.

Despite this progress, efficient long-sequence generation remains a significant challenge. In standard autoregressive decoding, each token must attend to the full KV cache, leading to frequent memory access and increased IO pressure. This bottleneck severely limits throughput, especially in long-context scenarios where memory access dominates latency.

\begin{wrapfigure}{r}{0.35\textwidth}
\setlength\intextsep{0pt}
\centering
\vspace{-1.2em}
\includegraphics[width=\linewidth]{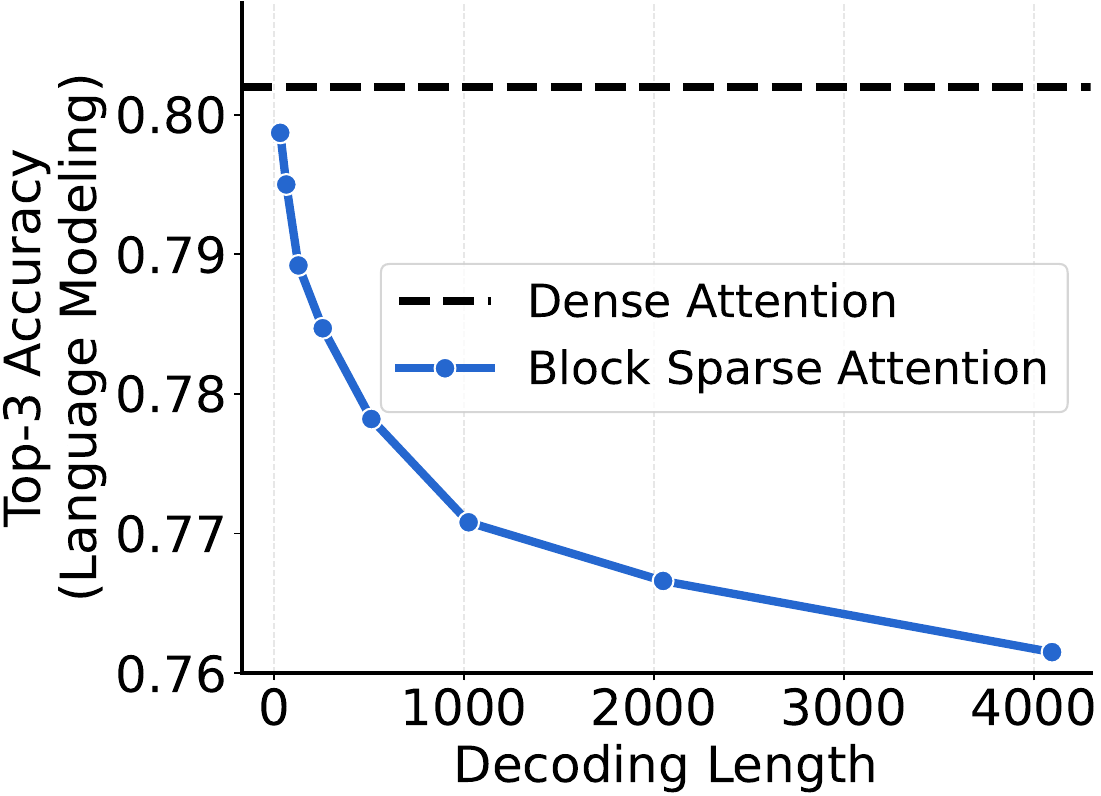}
\vspace{-1.5em}
\caption{Sparse decoding performance becomes worse with increasing decoding length due to error accumulation of KV cache.}
\vspace{-1em}
\label{fig:intro}
\end{wrapfigure}

Recent works~\cite{clusterkv, quest} used sparse decoding to alleviate this issue. These methods selectively attend to a subset of the context, achieving accuracy comparable to dense attention on long inputs while reducing computational cost. However, as shown in \Cref{fig:intro}, they often suffer from worse performance with increasing length. Since \textbf{computation errors accumulate in the KV cache during sparse decoding}, the attention computation will suffer from the misalignment between training and inference, contributing to performance degradation.

In this work, we propose Rectified Sparse Attention (\our{}), a simple yet effective approach that achieves near-lossless long-sequence generation quality while maintaining high inference efficiency. \our{} leverages block-sparse attention~\citep{quest} for fast retrieval and further improves memory efficiency by applying shared grouping~\citep{nsa}, allowing query heads to reuse attention patterns. To address the error accumulation issue, we introduce dense rectification, where the sparse KV cache is periodically refreshed with a parallel dense forward pass. This ensures that approximation errors are bounded within a constant range, preventing long-term degradation.

We conduct comprehensive experiments to demonstrate the effectiveness of \our{}. On math reasoning benchmarks, \our{} achieves strong test-time scaling and matches dense attention in long-sequence settings. In language modeling, \our{} significantly closes the quality gap between sparse and dense decoding. On the efficiency side, our approach yields up to 2.42$\times$ end-to-end speedup under INT4 decoding at 256K context length, showing strong practical utility for real-world deployment.

\section{Rectified Sparse Attention}
\begin{figure}[t]
\centering
\includegraphics[width=\textwidth]{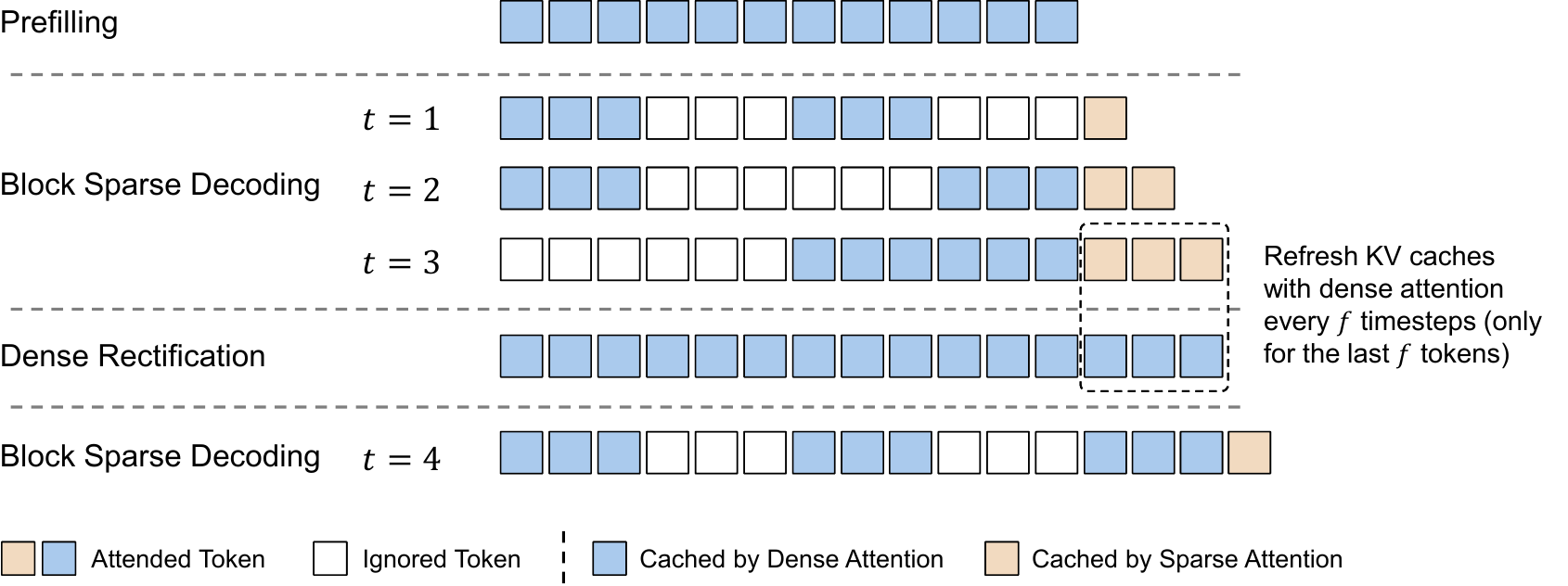}
\caption{Overview of \our{}. After completing the prefill stage, the model enters sparse decoding. Once the number of generated tokens reaches the rectification frequency, a rectification step is performed to construct a lossless compact KV cache, after which sparse decoding resumes.}
\label{fig:resa}
\end{figure}

\our{} primarily involves two alternating phases, sparse decoding and periodic rectification. During the decoding phase, we employ the group block sparse attention mechanism, which significantly reduces computational and memory overhead, enabling fast autoregressive inference.
During the rectification stage, the decoding tokens are forwarded in parallel to correct approximation errors in KV cache introduced by sparse decoding.
By alternating between sparse generation and dense rectification, \our{} enables scalable long-context inference while ensuring the generation quality.

\subsection{Group Block Sparse Attention}
\label{sec:gbsa}

Self-attention mechanisms are the core component of Transformer architectures, enabling each token to attend to all previous tokens. Formally, in Group-Query Attention (GQA)~\cite{gqa}, given a sequence of $n$ tokens, we compute the query $Q \in \mathbb{R}^{h\times g \times n \times d}$, key $K \in \mathbb{R}^{h\times n \times d}$, and value $V \in \mathbb{R}^{h\times n \times d}$ matrices through learned projections. The attention output is computed as:
\begin{equation}
\mathrm{Attention}(Q, K, V)_{ij} = \mathrm{softmax}\left(\frac{Q_{ij}K_{i}^\top}{\sqrt{d}}\right)V_{i}
\end{equation}
where $\mathrm{softmax}(\cdot)$ is applied along each query row. The pairwise computation requires $\mathcal{O}(n^2 d)$ operations, making standard attention prohibitively expensive for long-context inference.

We adopt a block-sparse attention design that selectively attends to a small number of relevant memory blocks rather than the entire context. Given the block size $b$ and block sparse mask $M\in \{0,1\}^{h\times n \times n/b}$, the block-sparse attention is computed as:
\begin{equation}
\mathrm{GBSA}(Q, K, V, M)_{ij} = \mathrm{softmax}\left(\frac{Q_{ij}K_{i}^\top\cdot\overline{M}_{i}}{\sqrt{d}}\right)V_{i},\ \overline{M}_{ijk}=M_{ij \lfloor k/b \rfloor}
\end{equation}

GBSA adopts a query-dependent sparsity pattern, where each query attends to a limited set of key blocks determined by $M$. Since each selected key block corresponds to a contiguous memory region in the KV cache, this design ensures both high performance and memory efficiency during inference.
Note that we further accelerate decoding by maintaining a shared sparse pattern within each GQA group~\cite{nsa}.

\begin{figure}[t]
    \centering
    \includegraphics[width=\textwidth]{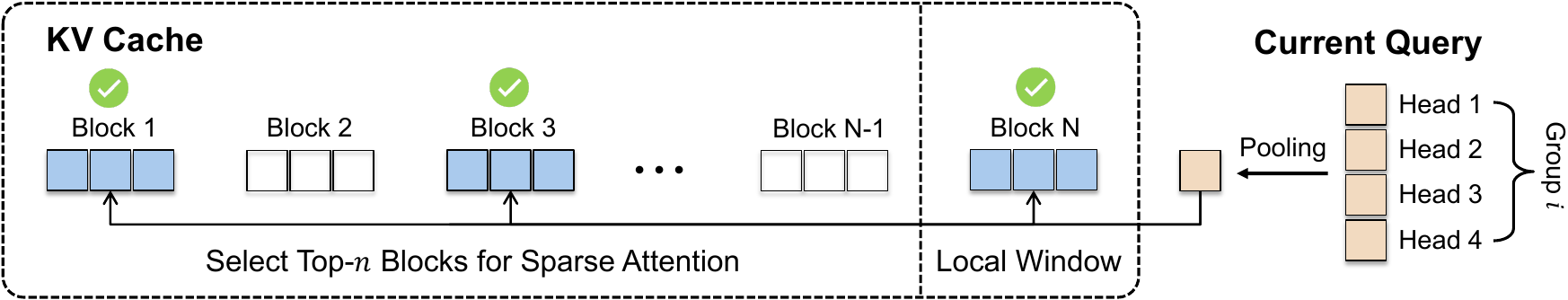}
    \caption{Overview of Group Block Sparse Attention. For each group of query heads, we perform average pooling and enforce the selection of the same KV blocks across all heads within the group.}
    \label{fig:blockselection}
\end{figure}

\paragraph{Block Representation}
Following the Quest algorithm~\citep{quest}, we represent the key-value memory using blocks to enable efficient retrieval. Specifically, given a key matrix $k \in \mathbb{R}^{n \times d}$, we partition it into non-overlapping blocks of size $b$, where each block contains $b$ consecutive tokens. For the $i$-th block, we compute two block descriptors:
\begin{equation}
k_{\text{block\_min},i} = \min(k_{ib:(i+1)b}),\ k_{\text{block\_max},i} = \max(k_{ib:(i+1)b})
\end{equation}
where $\min(\cdot)$ and $\max(\cdot)$ are applied element-wise across the block dimension.

Each block is thus summarized by a pair of vectors $(k_{\text{block min},i}, k_{\text{block max},i})$, which compactly describe the distribution range of keys within the block. This representation allows efficient approximate matching without exhaustively scanning all individual tokens. During decoding, newly generated keys can be incrementally incorporated by updating the block key statistics, enabling an online update mechanism without recomputing from scratch.  

Notably, the block representation is entirely training-free, relying solely on statistical descriptors. Our method remains compatible with more advanced block representation strategies, such as SeerAttention~\cite{seerattention}, where block keys are fine-tuned jointly with the model to achieve higher retrieval precision if needed.

\paragraph{Block Selection}
During decoding, given a pooling query $q \in \mathbb{R}^d$ for each GQA group and a set of block descriptors $\{(k_{\text{block\_min},i}, k_{\text{block\_max},i})\}_{i=1}^M$, we compute similarity scores following the Quest algorithm~\citep{quest}. Specifically, the score between the pooling query and block $i$ is calculated as:
\begin{equation}
\text{score}_i = \sum_{j=1}^d \max(q_j \times (k_{\text{block\_max},i})_j, \, q_j \times (k_{\text{block\_min},i})_j)
\end{equation}
where $q_j$ denotes the $j$-th dimension of the pooling query, and $(k_{\text{block min},i})_j$, $(k_{\text{block max},i})_j$ are the $j$-th dimensions of the minimum and maximum vectors of block $i$, respectively.

To select the attended blocks, we adopt a dynamic top-$n$ strategy. First, a fixed number of recent blocks, denoted as $n_{\text{local}}$, are always preserved by setting their scores to $+\infty$, ensuring that the latest context is available for local coherence. Second, we enforce a minimal block number $n_{\text{min}}$ to avoid significant performance degradation on short sequences. Finally, the value of $n$ is dynamically determined based on a active ratio $p$, following:
\begin{equation}
n = \max\left(n_{\text{min}}, \lceil M \times p \rceil\right),
\end{equation}
where $M$ is the total number of available memory blocks. Attention computation is restricted to the selected blocks, significantly reducing memory accesses while maintaining retrieval quality.

\subsection{Dense Rectification}

Transformer inference implicitly consists of two distinct phases: \textbf{context encoding}, realized through the construction of the KV cache, and \textbf{next-token prediction}, realized through the forward pass of the current token. While sparse attention effectively approximates the next-token prediction phase, it inevitably introduces errors. Crucially, these prediction errors accumulate in the KV cache during decoding, leading to compounding inaccuracies over long sequences.  
To mitigate this issue, we propose \textbf{Dense Rectification}, a lightweight mechanism that periodically refreshes the KV cache to maintain its quality. This design constrains error accumulation within a constant window size and enables efficient sparse decoding without compromising generation consistency.

\paragraph{Rectification Algorithm}
Given a rectification frequency $f$, we perform standard sparse decoding for up to $f$ tokens, appending newly generated tokens into the KV cache. After every $f$ token, we batch these recent tokens and re-encode them using dense attention to reconstruct an updated KV cache. This two-phase approach --- serial sparse decoding followed by parallel rectification --- ensures that errors introduced by approximate attention are corrected at regular intervals, keeping the memory quality close to that of dense decoding. Importantly, the rectification step amortizes efficiently over large batches, maintaining high throughput even when dense recomputation is involved.  
To maintain consistency, we also refresh the associated block keys during rectification. otherwise, the misalignment between the block keys and the updated KV cache would degrade subsequent sparse retrieval accuracy.


\paragraph{Compatibility with LLM Serving Systems}
Dense Rectification is naturally compatible with modern LLM serving optimizations such as continuous batching~\citep{orca} and chunked prefill~\citep{sarathi, fastgen}. Since rectification only requires periodic batched re-encoding, it seamlessly fits into systems that dynamically group decoding and prefill workloads to maximize GPU utilization. By maintaining a fixed rectification frequency per request, our method can operate within the batching and scheduling pipelines without introducing special synchronization barriers or inefficiencies.

\subsection{Decoding Procedure}
\begin{algorithm}[t]
\caption{Rectified Sparse Decoding}
\label{alg:context-rectification}
\begin{algorithmic}
\REQUIRE Initial prompts $\mathcal{P}$, model $\mathcal{M}$, rectification frequency $f$, maximum generation steps $T$
\ENSURE Generated tokens $\mathcal{G}$

\STATE Initialize KV cache $\mathcal{K}$ by $\texttt{Prefill}(\mathcal{P}, \mathcal{K})$
\STATE Initialize block key cache $\mathcal{B}$
\STATE Initialize output sequence $\mathcal{G} \leftarrow$ empty

\FOR{$i = 1$ \TO $T$}
    \STATE $t \leftarrow \texttt{SparseForward}(\mathcal{G}[i-1], \mathcal{K}, \mathcal{B})$
    \STATE Append $t$ to $\mathcal{G}$
    \STATE Update KV cache $\mathcal{K}$ with $t$
    \STATE Update block key cache $\mathcal{B}$ with $t$
    \IF{$i \bmod f = 0$}
        \STATE $\mathcal{K}, \mathcal{B} \leftarrow \texttt{DenseForward}(\mathcal{G}[i-f:i], \mathcal{K}, \mathcal{B})$
        \STATE Update block key cache $\mathcal{B}$
    \ENDIF
\ENDFOR

\end{algorithmic}
\end{algorithm}

Our decoding procedure alternates between sparse decoding and periodic rectification to achieve a balance between efficiency and generation quality. The process begins with a standard dense prefill phase, where the initial prompt is encoded into a complete key-value memory for subsequent decoding.
During the decoding phase, tokens are generated sequentially using sparse attention, which restricts memory access to a dynamically selected subset of context blocks. This enables fast autoregressive generation with reduced computational and memory costs.

To correct for approximation errors introduced by sparse attention, we periodically perform rectification. Specifically, after a fixed number of decoding steps, we batch the recently generated tokens and re-encode them using dense attention. This refreshes the key-value memory and ensures that accumulated errors are bounded within a constant window, maintaining memory quality close to dense decoding. The full decoding procedure is summarized in Algorithm~\ref{alg:context-rectification}.

The pipeline continues by alternating between sparse generation and rectification until the generation process completes. This design enables scalable long-context inference while preserving the consistency and reliability of the generated outputs.

\paragraph{Memory Access Analysis}
We further analyze the memory efficiency of the proposed decoding pipeline. In each sparse decoding step, the memory access consists of two parts: retrieving block keys for selection, proportional to $\operatorname{mem}(\text{KV cache}) / b$, and performing sparse attention, proportional to $\operatorname{mem}(\text{KV cache}) \times p$, where $b$ denotes the block size and $p$ denotes the sparsity ratio. In addition, for every $f$ step, a dense rectification is performed, whose amortized cost per step is $\operatorname{mem}(\text{KV cache}) / f$. Therefore, the average memory access per decoding step can be approximated as:
\begin{equation}
\operatorname{Avg}(\operatorname{mem}) = \operatorname{mem}(\text{KV cache}) \times \left( \frac{1}{b} + p + \frac{1}{f} \right).
\end{equation}
Compared to dense decoding, which requires accessing the entire KV cache at every step, our design achieves a theoretical memory access reduction factor of $\frac{1}{b} + p + \frac{1}{f}$.
By adjusting $b$, $p$, and $f$, the pipeline can flexibly trade-off between memory efficiency and generation fidelity.

\subsection{Kernel Implementation}

We develop a custom kernel optimized for the decoding phase, following a split-execution strategy similar to Flash Decoding and incorporating shared KV fetching techniques~\cite{nsa}. The key design principle is to assign each GQA group to an individual streaming multiprocessor (SM), ensuring efficient resource utilization and minimal inter-SM communication.

The decoding workload is $\mathrm{batch\_size}\times\mathrm{num\_kv\_heads}$. Given the total number of SMs available on the GPU, the workload is split accordingly to balance the computation between SMs. 
The splitting is performed at the level of block indices. For each decoding step, a batch of queries typically activates $k$ memory blocks. We partition these $k$ active blocks evenly across the available SMs, so that each SM is responsible for approximately $k/\text{split}$ blocks. Each SM independently fetches the required KV entries corresponding to its assigned blocks and performs sparse attention locally. The kernel implementation details are described in \Cref{appendix:gbsa}.

The design achieves high decoding throughput by minimizing memory contention, maximizing SM occupancy, and fully exploiting intra-GQA key sharing during sparse decoding.

\section{Experiments}
\subsection{Setup}
We evaluate \our{} from different perspectives. First, we make test-time scaling inference on math reasoning tasks~(\Cref{sec:tts}). Second, we simulate inference-time attention pattern on language modeling~(\Cref{sec:lm}). Third, we verify the effectiveness on retrieval~(\Cref{sec:needle}) tasks. Fourth, we analyze the inference advantages~(\Cref{sec:efficiency}, including kernel-level and end-to-end accelerations.

We choose Qwen2.5~\citep{qwen2.5}, a widely-used standard Transformer pre-trained model as evalutaion architectures. We apply \our{} on all of the layers, rather than skipping the first two layers in Quest~\citep{quest}. The block size is 16 and the minimal selected block number is $n_{\text{min}}=16, n_{\text{local}}=1$ to avoid performance degradation in short context. For longer sequences, the default sparsity ratio is $p=0.9$. The default rectification frequency is $f=32$.

\subsection{Long Reasoning}
\label{sec:tts}

We evaluate test-time scaling performance on math reasoning tasks. The validation dataets inclue Minerva Math~\citep{minerva}, Gaokao 2023 En~\citep{mario}, OlympiadBench~\citep{olympiadbench}, AIME24, and AMC23. We exclude some well-known math datasets such as GSM8K~\citep{gsm8k}, and MATH~\citep{math} since these datasets' average inference length is below 512. We choose DeepSeek-R1-Qwen-Distill 7B~\citep{deepseekr1} as the evaluation model. The number of attention head is 28 and kv head is 4. The hidden size is 3584 and the number of layers is 28.

The results in \Cref{tab:math} show that while \our{} achieves performance comparable to the dense baseline, Sparse Decoding alone consistently underperforms. \our{} maintains near-lossless performance in long-context reasoning tasks, whereas Sparse Decoding leads to performance degradation as decoding progresses. Additionally, manually enforcing dense layers for the first two layers does not result in a significant improvement in math-reasoning tasks.

\begin{table}[t]
\centering
\resizebox{\linewidth}{!}{%
\begin{tabular}{lcccccc}
\toprule
 & \textbf{Minerva} & \textbf{Gaokao2023En} & \textbf{OlympiadBench} & \textbf{AIME24} & \textbf{AMC23} & \textbf{Avg} \\
\midrule
\multicolumn{7}{l}{\textit{R1-Qwen-Distill 1.5B}} \\
Dense          & 28.7 & 71.6 & 40.8 & 27.4 & 65.6 & 46.82 \\
\noalign{\vspace{1pt}} 
\cdashline{1-7}
\noalign{\vspace{2pt}} 
Sparse & \textbf{29.0} & 67.9 & 38.7 & 21.3 & 60.6 & 43.50 \\
\our{} & 28.1 & \textbf{71.8} & \textbf{39.5} & \textbf{23.0} & \textbf{65.4} & 45.56 \\
\midrule
Avg Length & 6390.8 & 4915.8 & 8991.6 & 12126.4 & 7866.4 & 8058.2 \\
\midrule
\midrule
\multicolumn{7}{l}{\textit{R1-Qwen-Distill 7B}} \\
Dense         & 40.4 & 73.8 & 52.3 & 48.1 & 89.0 & 60.72 \\
\noalign{\vspace{1pt}} 
\cdashline{1-7}
\noalign{\vspace{2pt}} 
Sparse    & 38.1 & 72.9  & 48.4 & 46.1  & 83.1  & 57.72 \\
Sparse$_{\mathrm{dense}2}$    & 37.9 & 72.5  & 48.8 & 44.6  & 83.1  & 57.38 \\
\our{}   & \textbf{39.7} & \textbf{73.5}  & \textbf{52.3} & \textbf{51.1}  & \textbf{86.0}  & 60.52 \\
\midrule
Avg Length        & 4018.7 & 2889.9 & 7520.0 & 10474.5 & 5732.2 & 6127.1 \\
\bottomrule
\end{tabular}
}
\caption{Performance comparison on math reasoning tasks. While simple sparse decoding methods show a gap with dense decoding, \our{} achieves near lossless long-sequence generation. }
\label{tab:math}
\end{table}

\subsection{Language Modeling}
\label{sec:lm}

\begin{figure}[t]
    \centering
    \begin{minipage}[t]{0.45\textwidth}
        \centering
        \includegraphics[width=\linewidth]{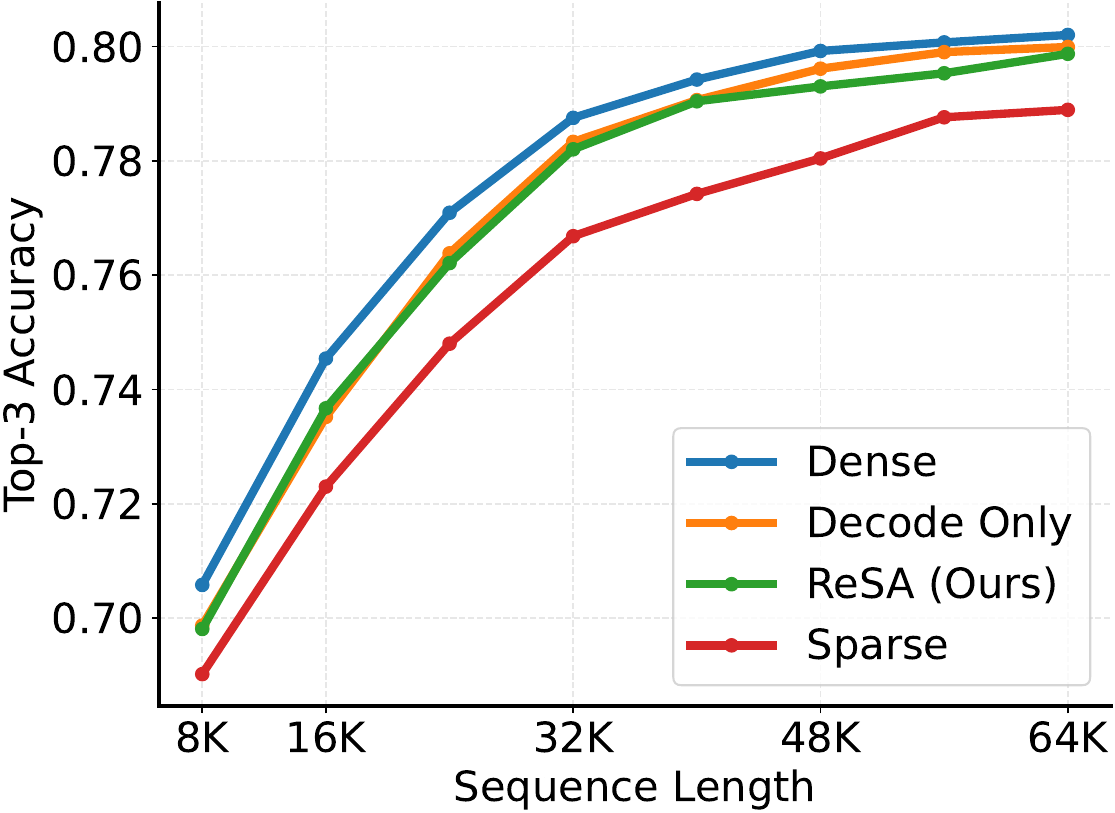}
        \caption{Top-3 next-token prediction accuracy with different rectification frequency.}
        \label{fig:ppl_main}
    \end{minipage}
    \hfill
    \begin{minipage}[t]{0.45\textwidth}
        \centering
        \includegraphics[width=\linewidth]{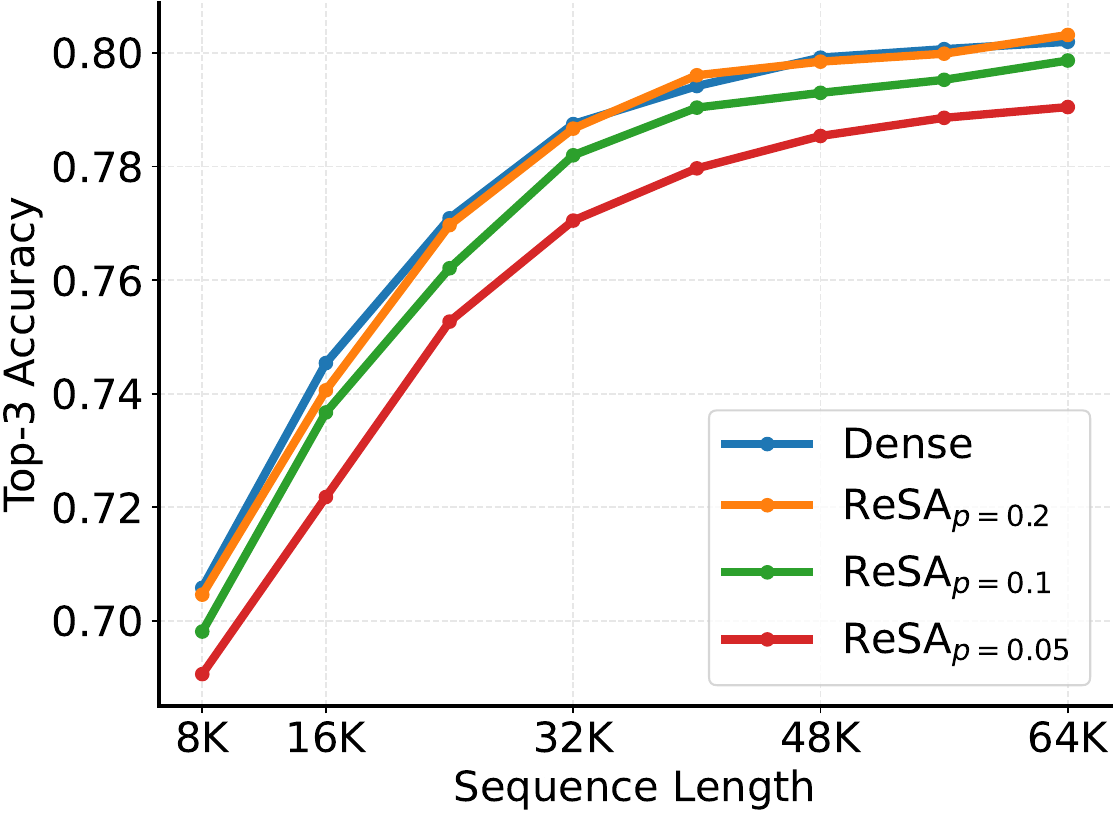}
        \caption{Top-3 next-token prediction accuracy with different sparsity ratio.}
        \label{fig:ppl_spa}
    \end{minipage}
\end{figure}

Se evaluate language modeling performance under simulated sparse decoding patterns. Specifically, we divide each input sequence into two parts. Given a total sequence length $L$, we split it into a prefix of length $L - x$ and a suffix of length $x$. The prefix is processed using dense attention, while the suffix uses sparse attention. Here, $x$ effectively controls the rectification frequency. When $x = L$, it corresponds to the sparse decoding baseline, where no rectifying is performed and the entire sequence is encoded using sparse attention.

We conduct our experiments using long-sequence book data. For each target sequence length, we use the same data and truncate from the left to ensure that the prediction tokens are perfectly aligned across all settings. We report the top-3 accuracy computed over the final 32 tokens of each sequence to focus on the model's performance in the later decoding stages.

As shown in Figure~\ref{fig:ppl_main}, we compare the impact of different rectification frequencies on model perplexity. The setting labeled \textit{Decode Only} corresponds to the case where all KV cache entries are generated using dense attention, and sparse attention is only used for decoding. This serves as the upper bound for \our{}. We observe that \our{} significantly reduces the performance gap between dense and sparse decoding. Notably, when $x = 32$, the model's performance almost approaches the upper bound, demonstrating the effectiveness of rectification in mitigating the error accumulation issue inherent in sparse decoding.

In Figure~\ref{fig:ppl_spa}, we further examine the effect of different sparsity ratios under a fixed rectification frequency of $x = 32$. We find that there is a noticeable performance gap between the $p=0.98$ and $p=0.95$. Although $p=0.8$ sparsity achieves perplexity comparable to the dense setting, we adopt $p=0.9$ as the default due to its better trade-off between performance and efficiency. Additionally, since effective block selection strategies can lead to higher achievable sparsity, our method can be further combined with advanced attention selection mechanisms such as SeerAttention~\cite{seerattention} to enhance runtime efficiency.

\subsection{Long-Sequence Retrieval}
\label{sec:needle}

\begin{table}[t]
\centering
\resizebox{\textwidth}{!}{%
\begin{tabular}{@{}lccccccccc@{}}
\toprule
\textbf{Setting} & \textbf{QA} & \textbf{MultiQuery} & \textbf{FWE} & \textbf{VT} & \textbf{MultiKey} & \textbf{MultiValue} & \textbf{CWE} & \textbf{Single} & \textbf{Avg} \\
\midrule
Dense & 0.563 & 0.211 & 0.833 & 0.719 & 0.688 & 0.246 & 0.134 & 1.000 & 0.549 \\
\midrule
\our{}$_{p=0.95}$ & 0.500 & 0.180 & 0.740 & 0.719 & 0.750 & 0.238 & 0.125 & 1.000 & 0.531 \\
\our{}$_{p=0.9}$ & 0.625 & 0.203 & 0.760 & 0.719 & 0.750 & 0.234 & 0.178 & 1.000 & 0.559 \\
\our{}$_{p=0.8}$ & 0.594 & 0.195 & 0.771 & 0.719 & 0.719 & 0.246 & 0.175 & 1.000 & 0.552 \\
\bottomrule
\end{tabular}
}
\caption{RULER benchmarks under different sparsity ratios. Dense represents the fully-attended baseline, while \our{}$_{p=x}$ denotes our method with sparsity level $x$.}
\label{tab:ruler}
\end{table}

We conduct experiments on the RULER benchmark to further evaluate the impact of different sparsity levels. Unlike the long-sequence generation tasks, where rectification plays a critical role in mitigating cumulative error, the RULER benchmark focuses on relatively short output sequences. As a result, the final accuracy is primarily determined by the quality of the sparse attention estimation.

Results are presented in Table~\ref{tab:ruler}.
We observe that as the sparsity ratio increases from $p=0.95$ to $p=0.9$, there is a consistent improvement in average accuracy, with \our{}$_{p=0.9}$ achieving comparable performance to the dense baseline (0.559 vs. 0.549). The performance under $p=0.8$ remains similar to that under $p=0.9$, indicating that moderate increases in sparsity do not substantially degrade accuracy in short-generation settings. Considering that a lower sparsity ratio generally leads to faster inference, \our{}$_{p=0.9}$ represents a better trade-off between performance and efficiency on the RULER benchmark.

\subsection{Inference Efficiency}
\label{sec:efficiency}

We evaluate the efficiency of \our{} on standard GPU hardware. Specifically, we use Qwen-2.5 7B as the evaluation model and conduct all experiments on NVIDIA A100-80G GPUs. The primary baseline is FlashAttention, a highly optimized dense attention implementation. To ensure a fair comparison and prevent memory overflow issues caused by excessively large KV caches during long-sequence evaluation, we adopt a shared KV cache strategy across all layers during efficiency measurements. The batch size is fixed at 8 by default throughout all experiments.

For latency measurement, we report the CUDA kernel execution time, excluding CPU-side scheduling overhead. This setup more accurately reflects the real-world inference scenario, as the CPU overhead can be effectively optimized away through techniques such as CUDA graph capture.

\subsubsection{Attention Efficiency}

\begin{figure}[t]
\centering
\includegraphics[width=\textwidth]{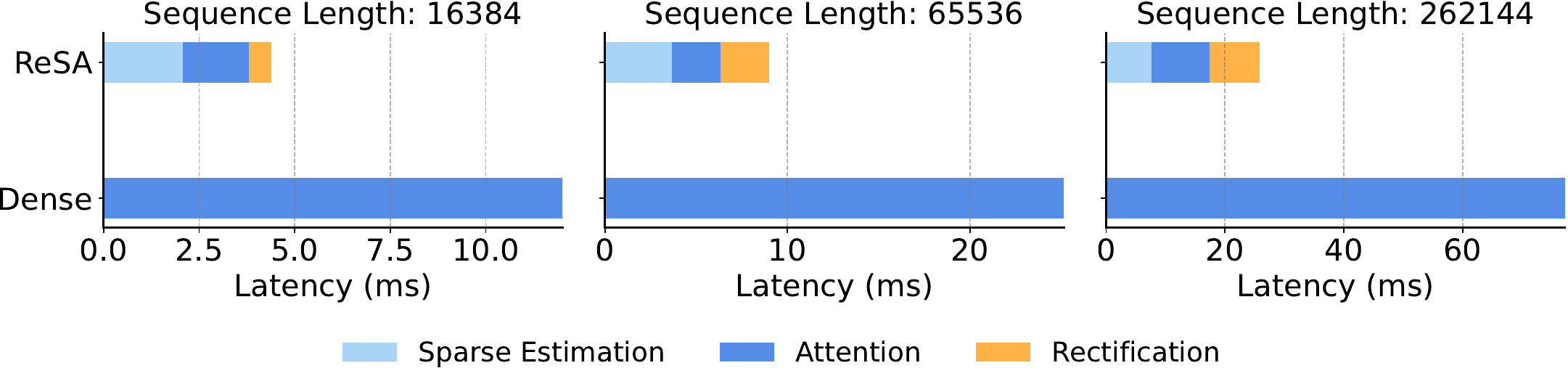}
\caption{Kernel-level latency breakdown across different sequence lengths. While Sparse Decoding achieves effective acceleration, rectification only requires a small additional overhead.}
\label{fig:latency}
\end{figure}

Figure~\ref{fig:latency} shows the detailed latency breakdown across different sequence lengths ($16$k, $64$k, and $256$k tokens). We compare \our{}, and dense attention under the same settings. The latency is decomposed into three parts: sparse estimation, attention computation, and rectification overhead.

Compared to dense attention, \our{} significantly reduces the total latency, especially at longer sequence lengths. As the sequence grows, dense attention exhibits longer latency with increasing context length, leading to substantial latency increase, while \our{} maintains much flatter scaling due to its sparsified attention computation.

Moreover, sparse estimation and attention computation consume comparable amounts of time, because the memory access pattern for sparse estimation scales with $\text{mem}(\text{KV cache})/\text{block}$, while for attention it scales with $\text{mem}(\text{KV cache}) \times p$. Given our experimental settings ($\text{block}=16$, $p=0.9$), both operations operate on similar memory volumes. Notably, under fixed block size, further increasing the sparsity ratio can not bring significant speed-up. 

The overhead of rectification is relatively small compared with sparse decoding part. Specifically, the rectification module accounts for up to 32.7\% of the total attention-related latency at 256k lengths, while at 64k, this proportion drops to 28.9\%. When the sequence length is scaling, the latency ratio will converge to the memory access ratio $1/f$. These results indicate that while sparse estimation and attention computation remain efficient, the rectification does not bring big overhead.

\subsubsection{End-to-End Efficiency}

\begin{figure}[b]
    \centering
    \begin{minipage}{0.48\textwidth}
        \centering
        \includegraphics[width=\linewidth]{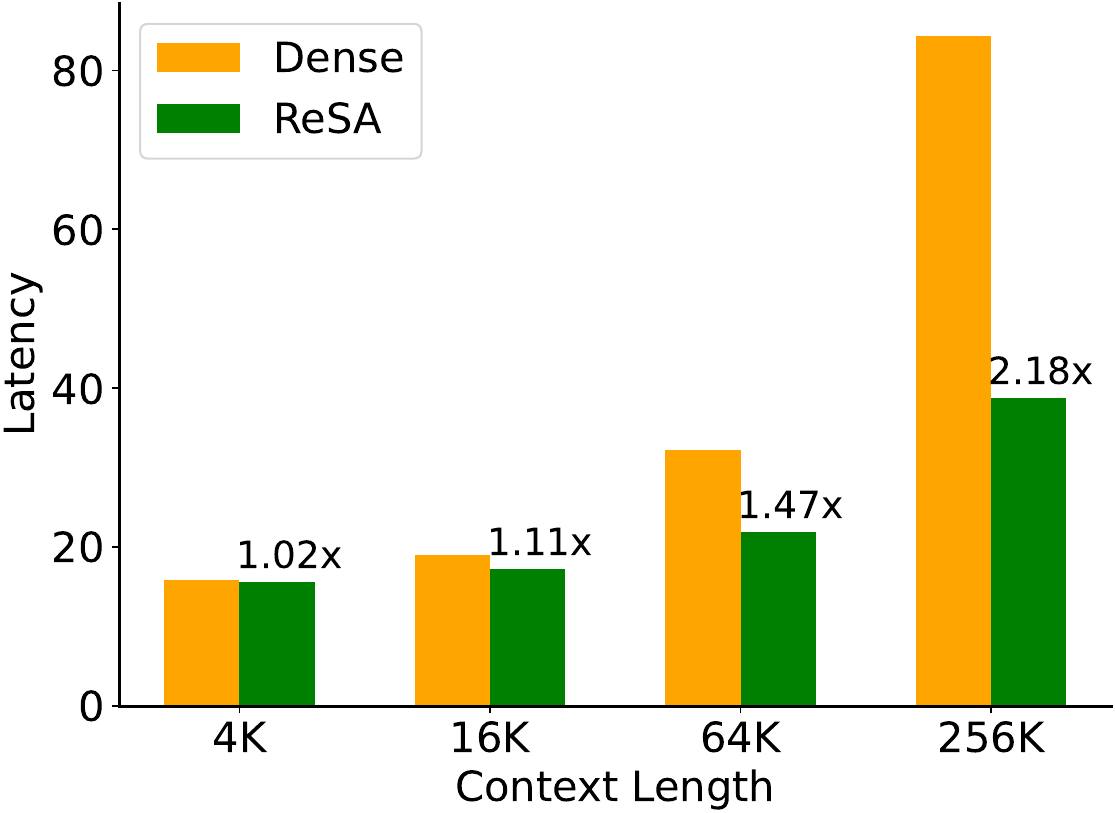}
        \caption{End-to-end latency with FP16.}
        \label{fig:e2e_fp16}
    \end{minipage}
    \hfill
    \begin{minipage}{0.48\textwidth}
        \centering
        \includegraphics[width=\linewidth]{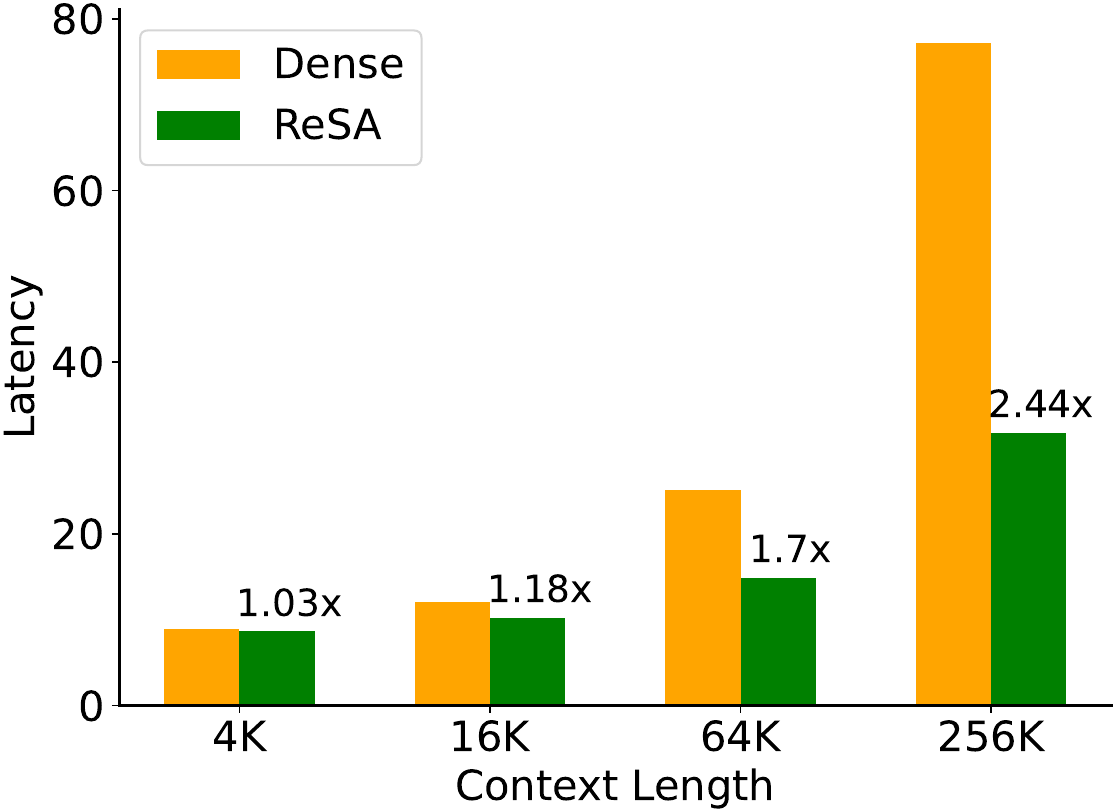}
        \caption{End-to-end latency with INT4.}
        \label{fig:e2e_int4}
    \end{minipage}
\end{figure}

We further evaluate the end-to-end throughput of \our{} in both FP16 and INT4 precision settings. For the INT4 experiments, we leverage the Marlin kernel~\cite{frantar2024marlin} for low-bit matmul. The matmul weight is quantized with group-wise scaling. The group size is 128.

Figure~\ref{fig:e2e_fp16} and Figure~\ref{fig:e2e_int4} report the throughput across different context lengths (4K, 16K, 64K, and 256K tokens) under FP16 and INT4 settings, respectively. Consistent with the kernel-level results, \our{} significantly improves the overall throughput as the sequence length grows, achieving up to $2.28\times$ speedup over dense attention in FP16 and $2.44\times$ in INT4 at 256K context length.

Notably, the benefits of \our{} become more prominent at longer sequences due to the quadratic scaling bottleneck of dense attention, while the overhead of sparse estimation and rectification remains modest even under quantized inference. These results demonstrate that \our{} is highly effective in improving real-world end-to-end generation speed across different precision levels.

\subsection{Ablation Studies}
\label{sec:ablation}

\begin{figure}[t]
\centering
\includegraphics[width=\textwidth]{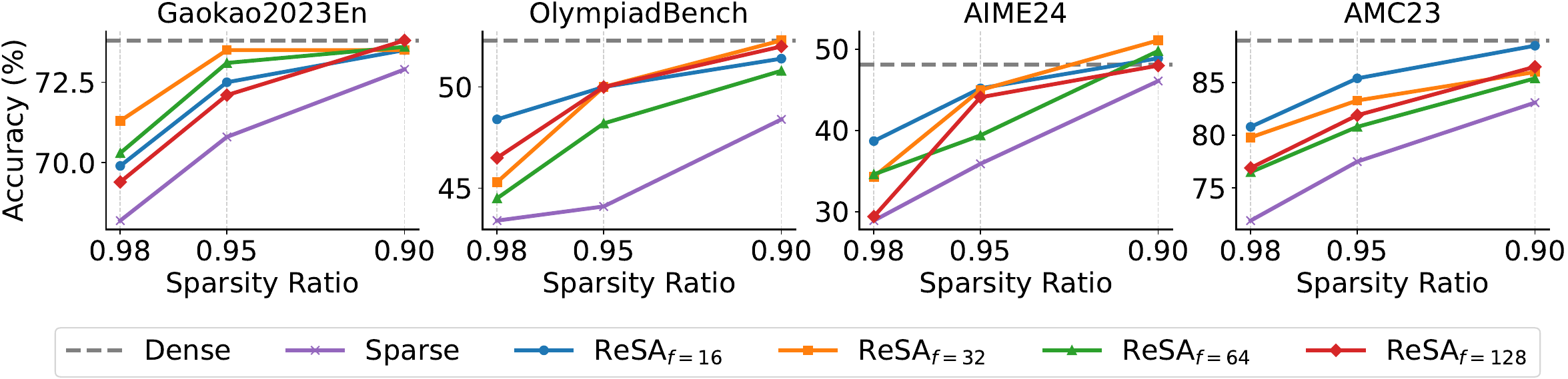}
\vspace{-1em}
\caption{Ablation studies on different rectification frequencies $f$ and sparsity ratios $p$ across five math reasoning benchmarks. \our{} consistently improves over the sparse baseline. Frequencies $f=32$ or $f=64$ achieve the best trade-off between performance and overhead.}
\label{fig:ablation}
\vspace{-1em}
\end{figure}

We conduct ablation studies to examine the effect of rectification frequency and sparsity ratio on performance. As shown in \Cref{fig:ablation}, we evaluate \our{} across five math reasoning benchmarks under varying sparsity levels ($p \in \{0.9, 0.95, 0.98\}$) and rectification frequencies ($f \in \{16, 32, 64, 128\}$).

Compared to the sparse decoding baseline, \our{} consistently outperforms the baseline across all sparsity levels. Notably, when the attention computation ratio is reduced to 0.1, \our{} achieves accuracy that is remarkably close to the dense decoding upper bound. This demonstrates that \our{} effectively mitigates the quality drop typically associated with sparse decoding while maintaining high computational efficiency.

Among the frequencies, $f=32$ achieves accuracy close to the dense baseline on most datasets, striking a favorable balance between quality and efficiency. While $f=16$ offers marginal gains, it incurs higher rectification overhead and is therefore less practical. Notably, even with $f=128$, a large portion of the performance gain is retained, highlighting the robustness of the rectification mechanism under infrequent updates.

\section{Related Work}
\label{sec:relatedwork}

\textbf{Sparse Attention}~~Recent efforts in sparse decoding for large language models can be broadly categorized into training-free and training-aware approaches. Training-free methods enhance inference efficiency without substantial retraining. Quest~\citep{quest} and InfLLM~\citep{infllm} both adopt query-aware block-sparse attention, selectively retrieving critical memory blocks based on query relevance. MagicPig~\citep{magicpig} and ClusterKV (Tactic)~\citep{clusterkv} employ similarity-based techniques, using hashing or clustering to approximate attention relevance. In contrast, training-aware architectures such as NSA~\citep{nsa} and MoBA~\citep{moba} integrate sparsity into model design, aligning structures with hardware during pretraining. Our method complements training-free sparse attention by improving memory quality through lightweight rectification, avoiding the high retraining cost required by training-aware approaches.

\textbf{Speculative Decoding}~~Speculative decoding~\citep{specdecoding} accelerates generation by drafting multiple tokens and verifying them with the target model. Methods like Medusa~\citep{medusa} and EAGLE~\citep{eagle} reuse the target model's hidden states for drafting. TriForce~\citep{triforce} and MagicDec~\citep{magicdec} propose self-speculation, using the model's own sparse KV cache for drafting and a dense cache for verification. While sharing similar compute characteristics with sparse KV-based self-speculation, \our{} avoids per-token accept/reject decisions and resampling overhead. In Appendix~\ref{appendix:self-speculation}, we compare \our{} and self-speculation in detail.

\section{Conclusion}

In this paper, we introduced Rectified Sparse Attention, a simple yet effective method for efficient long-sequence generation. \our{} combines group block sparse attention for decoding latency, and dense rectification to bound error accumulation. Extensive experiments on math reasoning and language modeling tasks demonstrate that \our{} achieves near-lossless performance compared to dense decoding, delivering up to 2.42$\times$ inference speedup at 256K context length. These results highlight \our{}'s practical effectiveness in long-context language model deployment.

\ifpreprintversion
\section*{Acknowledgments}

We would like to thank Lei Wang and Yu Cheng for their valuable help implementing group block sparse attention with the \href{https://github.com/tile-ai/tilelang}{TileLang} library.
\else

\fi

\bibliographystyle{plainnat}
\bibliography{arch}


\newpage
\appendix

\section{Pseudo Code of Flash Decoding Kernel}
\label{appendix:gbsa}

The proposed group block sparse attention (\Cref{sec:gbsa}) can be easily integrated into the Flash Decoding~\cite{flashdec} kernel implementation.
The modified parts are highlighted as follows.

\begin{algorithm}[htb]
\caption{Flash Decoding with Block-Sparse Attention}
\label{alg:flash-decoding}
\begin{algorithmic}[1]
\REQUIRE Queries $Q$, Keys $K$, Values $V$, $\mathrm{block\_indices}$
\ENSURE Attention outputs $Out_{partial}, logsum_{partial}, Out$
\FOR{Grid indexed by $(\mathrm{num\_splits}, \mathrm{num\_kv\_heads}, \mathrm{batch\_size})$}
    \STATE \textcolor{linkColor}{Load query vectors $q$ in a GQA group}
    \STATE \textcolor{linkColor}{Compute $\mathrm{partial\_block\_indices}$ with $\mathrm{block\_indices}$ and $\mathrm{num\_splits}$}
    \STATE Initialize accumulators: $m_i \gets -\infty$, $l_i \gets 1.0$, $acc \gets 0$
    \FOR{\textcolor{linkColor}{$\mathrm{block\_id}$ in $\mathrm{partial\_block\_indices}$}}
        \STATE Load keys $k$ and values $v$ from KV cache in \textcolor{linkColor}{block $\mathrm{block\_id}$}
        \STATE Compute scaled attention scores $qk \gets (qk) \times sm\_scale$
        \STATE Apply masking to invalid positions ($qk \gets -1e6$)
        \STATE Compute and update $m_i, l_i, acc$
    \ENDFOR
    \STATE Store partial logsum and attention outputs into $\mathrm{logsum}_\mathrm{partial}, \mathrm{Out}_\mathrm{partial}$
\ENDFOR
\STATE Combine different splits $\mathrm{Combine}(\mathrm{logsum}_\mathrm{partial}, \mathrm{Out}_\mathrm{partial}, \mathrm{Out})$
\STATE \textbf{return} Attention output tensor $\mathrm{Out}$
\end{algorithmic}
\end{algorithm}

\section{Comparison with Self-Speculation}
\label{appendix:self-speculation}

As discussed in Section~\ref{sec:relatedwork}, \our{} shares similar computational characteristics with sparse KV cache-based self-speculation. The rectification phase in \our{} resembles the verification phase used in self-speculative methods. However, unlike these methods, \our{} does not rely on output logits to make per-token accept / reject decisions. This design choice is motivated by the observation that, when sparse attention achieves high generation quality, this kind of token-wise strict verification can significantly increase latency without providing proportionate accuracy gains.

To validate this, we compare \our{} and sparse KV-based self-speculation on mathematical reasoning tasks. We set the speculation length to 16, meaning that the model drafts 16 tokens using the sparse KV cache. Similarly, we set \our{}'s rectification frequency to 16. Across all tasks, \our{} achieves nearly 2× speedup over self-speculation while maintaining comparable accuracy. This is because, in each verification step of speculative decoding, only about 8 tokens are typically accepted—effectively halving the generation rate compared to \our{}. Although this strict verification ensures that speculative decoding matches the accuracy of dense attention, we have previously shown that \our{} also approaches the accuracy of dense attention. Therefore, we believe that the marginal accuracy gains of speculative decoding do not justify its substantial latency overhead.

\begin{table}[h]
\centering
\resizebox{0.8\textwidth}{!}{%
\begin{tabular}{lcc}
\toprule
\textbf{Task} & \textbf{Sparse KV Self-Spec.} & \textbf{Rectified Sparse Attention} \\
\midrule
Minerva & 1× & 1.93× \\
Gaokao2023En & 1× & 1.87× \\
OlympiadBench & 1× & 1.98× \\
AIME24 & 1× & 1.96× \\
AMC23 & 1× & 1.86× \\
\midrule
\textbf{Average} & 1× & 1.92× \\
\bottomrule
\end{tabular}
}
\vspace{1em}
\caption{Decoding speedup comparison. We set the throughput of self-speculation as baseline. \our{} achieves larger speedup compared with sparse self-speculative decoding.}
\label{tab:math_speedup_comparison}
\end{table}

\ifpreprintversion
\fi

\end{document}